\pdfoutput=1

\documentclass[11pt]{article}

\usepackage[final]{coling}

\usepackage{times}
\usepackage{latexsym}

\usepackage[T1]{fontenc}

\usepackage[utf8]{inputenc}

\usepackage{microtype}

\usepackage{inconsolata}

\usepackage{graphicx}
\usepackage{amsmath}
\usepackage{booktabs}
\usepackage{multirow}
\usepackage{makecell}
\usepackage{tcolorbox}

\title{Multi-Facet Counterfactual Learning for Content Quality Evaluation}

\author{
  Jiasheng Zheng${}^{1,2}$,
  Hongyu Lin${}^{1}$,
  Boxi Cao${}^{1,2}$,
  \\
  \textbf{Meng Liao${}^{3}$},
  \textbf{Yaojie Lu${}^{1}$},
  \textbf{Xianpei Han${}^{1}$},
  \textbf{Le Sun${}^{1}$}
  \\
  ${}^{1}$Chinese Information Processing Laboratory, Institute of Software \\
  Chinese Academy of Sciences, Beijing, China \\
  ${}^{2}$University of Chinese Academy of Sciences, Beijing, China \\
  ${}^{3}$Data Quality Team, WeChat, Tencent Inc., China \\
 {\tt \{zhengjiasheng2022,hongyu,boxi2020,luyaojie,xianpei,sunle\}@iscas.ac.cn} \\
 {\tt maricoliao@tencent.com }
}

\begin{document}
\maketitle
\begin{abstract}

Evaluating the quality of documents is essential for filtering valuable content from the current massive amount of information.
Conventional approaches typically rely on a single score as a supervision signal for training content quality evaluators, which is inadequate to differentiate documents with quality variations across multiple facets. 
In this paper, we propose \textbf{M}ulti-facet c\textbf{O}unterfactual \textbf{LE}arning (MOLE), a framework for efficiently constructing evaluators that perceive multiple facets of content quality evaluation. Given a specific scenario, we prompt large language models to generate counterfactual content that exhibits variations in critical quality facets compared to the original document.
Furthermore, we leverage a joint training strategy based on contrastive learning and supervised learning to enable the evaluator to distinguish between different quality facets, resulting in more accurate predictions of content quality scores. 
Experimental results on 2 datasets across different scenarios demonstrate that our proposed MOLE framework effectively improves the correlation of document content quality evaluations with human judgments, which serve as a valuable toolkit for effective information acquisition.

\end{abstract}

\section{Introduction}

Evaluating document quality is crucial for humans to efficiently acquire and filter valuable content from massive amounts of information. With the explosive growth of information, particularly the widespread dissemination of AI-generated content~\citep{cao2023comprehensive, wu2023ai}, human-based evaluation methods are proving inadequate when faced with large-scale data~\citep{li2024leveraging}. To this end, an increasing number of researchers are focusing on leveraging machine learning techniques as a substitute for human evaluation in assessing the quality of document content. Typically, they begin by collecting human-labeled document datasets with either discrete ratings or continuous score annotations, and then employ supervised learning techniques to train evaluators.

\begin{figure}[t!]
  \centering
  \setlength{\abovecaptionskip}{0.2cm}
  \setlength{\belowcaptionskip}{-0.7cm}
  \includegraphics[width=\columnwidth]{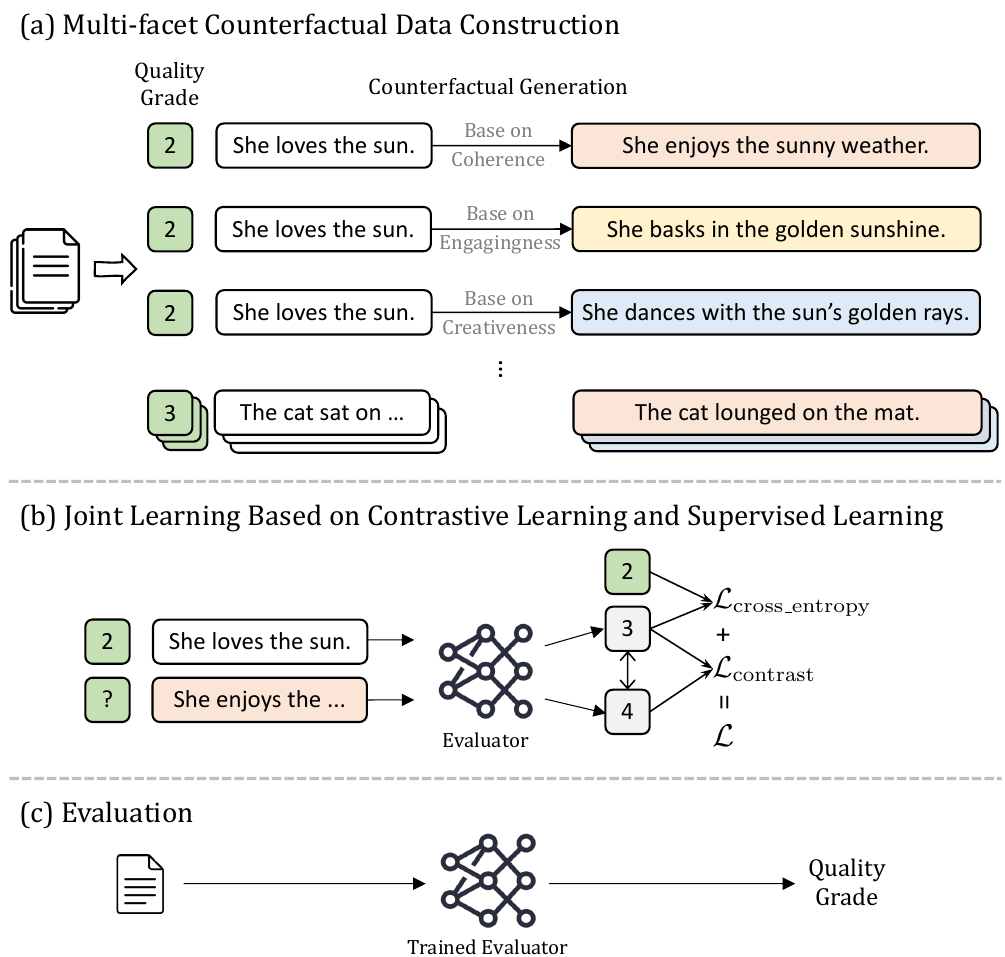}
  \caption{Overview of the MOLE framework: By constructing documents with counterfactual content under multiple quality facets and pairing them with the original documents, the framework facilitates contrastive learning, enabling the evaluator to better perceive different quality facets.}
  \label{fig:head}
\end{figure}

However, relying solely on such a scalar value is insufficient for the evaluator to learn how to differentiate documents with quality variations across multiple facets.
In particular, the quality of a document's content inherently encompasses multiple aspects, such as informativeness~\citep{conf/sigdial/MehriE20, yuan2021bartscore}, coherence~\citep{conf/acl/LiuF023, conf/emnlp/MaimonT23}, and engagingness~\citep{conf/acl/KielaWZDUS18, gopalakrishnan2023topical}. A dataset that provides only an overall quality rating or score not only fails to deliver detailed insights into content quality but is also likely to overlook the various facets of content quality that need to be considered.
Additionally, the definition and standards of document content quality vary across different application scenarios and requirements. This variation makes it challenging to share supervised learning signals across different contexts. 
Furthermore, manually creating specialized supervised datasets for each application scenario is both costly and difficult to adapt quickly to changing needs. 
Therefore, the core challenge in building content quality evaluation systems lies in how to efficiently and cost-effectively generate learning signals based on the specific requirements of a given scenario.

In this paper, we propose a novel framework named \textbf{M}ulti-facet c\textbf{O}unterfactual \textbf{LE}arning (MOLE). 
Given a specific evaluation scenario, MOLE prompts large language models to generate contrastive data with counterfactual content for annotated training documents across various quality facets, thereby introducing differences in specific quality facets.
Then, through joint learning based on contrastive learning and supervised learning, MOLE enhances the content quality evaluator’s ability to perceive different quality facets, achieving a higher level of consistency with human judgment.
Furthermore, MOLE can be quickly adapted to new scenarios based on varying requirements.

To verify the effectiveness of our framework, we conducted experiments on 2 document quality evaluation scenarios, which include Web-Article Quality Evaluation and ASAP datasets.
The experimental results show that MOLE significantly improves the correlation of the evaluator with human judgments, which can provide a valuable evaluator that covers the necessary multiple facets of quality.
Moreover, detailed analysis demonstrates that MOLE can improve the evaluator's assessment capabilities under extreme conditions and assist in comprehensively perceiving fine-grained quality dimensions.

\section{Multi-Facet Counterfactual Learning}

To enhance the consistency between evaluators' assessments and human judgments while improving their ability to perceive various quality facets, we propose the \textbf{MOLE} framework. 
Firstly, through multi-faceted counterfactual data generation, we leverage large language models (LLMs) to generate counterfactual content and construct a contrastive dataset that covers multiple quality facets, as detailed in Section~\ref{sec:MFCL_1}. Secondly, we employ joint learning based on contrastive learning and supervised learning, which allows the evaluators to better learn patterns that differentiate between various quality facets from the constructed contrastive dataset, as detailed in Section~\ref{sec:MFCL_2}.

\subsection{Multi-faceted Counterfactual Data Generation}

\label{sec:MFCL_1}

The goal of our multi-faceted counterfactual data generation is to efficiently construct contrastive datasets containing multi-faceted learning signals. First, given a specific scenario, we select the critical quality facets relevant to the evaluation task. 
Drawing inspiration from prior research on content quality evaluation in single-dimensional and single-scenario settings~\citep{conf/aaai/LiaoXLW21, conf/emnlp/LiWW22, conf/coling/XieCKZQ22}, we hypothesize that a high-quality general-purpose document demonstrates significant strengths in coherence, usefulness, creativeness, informativeness, and engagingness. These constitute our set of quality facets, with detailed definitions available in Appendix~\ref{sec:dimensions}.

Furthermore, using the defined set of quality facets, we begin with a dataset that includes overall quality scores to ensure the effectiveness of the constructed learning signals.
We then prompt an LLM\footnote{We use GPT-3.5-Turbo~\citep{chatgpt} in this paper to balance data quality and budget.} to identify issues related to a specific quality facet within a document, using its overall quality score as a reference.
Subsequently, based on the identified quality issues, the same LLM is directed to rewrite the document to address these concerns as effectively as possible. This process generates a pair of documents that differ in quality regarding that specific facet, consisting of the original document and a rewritten version featuring counterfactual content.
To generate multi-faceted learning signals, each document is randomly assigned a quality facet from the predefined set. The prompt templates are provided in Appendices~\ref{sec:prompts}.

\subsection{Joint Learning Based on Contrastive Learning and Supervised Learning}

\label{sec:MFCL_2}

After undergoing meticulous pre-training and alignment, LLMs exhibit a tendency to align with human values \citep{zhao2023survey}. Consequently, we believe that well-aligned LLMs hold significant potential to effectively learn from the constructed dataset covering multiple quality facets. To optimize the evaluation capabilities of LLMs and leverage their instruction-following ability, we frame the task of content quality evaluation as a question-answer task. By providing appropriate instructions, we guide the model to perform evaluations more effectively. The process can be formulated as $\mathbf{s}_i=\mathrm{LLMDecoder}(\mathbf{s}_{i-1}|\mathbf{x})$, 
where the LLM generates a sentence $\mathbf{s}$ of length $k$ that articulates the final evaluation in natural language.

To enable the evaluator to differentiate between various quality facets while ensuring accurate prediction of quality scores, we design a joint learning strategy based on contrastive learning and supervised learning.
To further activate the foundational capabilities of LLM-based evaluators, we make corresponding adaptations in both the model architecture and the format of the evaluation signals.
In terms of model architecture, to avoid disrupting the model's output vocabulary distribution, we introduce an additional linear layer before the model's decoding layer. This allows contrastive learning to be conducted based on the transformed hidden representations of the LLM.
For the evaluation signal format, we choose a 5-point Likert scale to help the model more effectively differentiate between different quality levels. We believe this setting would guide the model to learn to differentiate between fixed quality levels, rather than struggling with fine-grained comparisons between documents with very close quality, thereby reducing the difficulty of model training to some extent.
Ultimately, we design the loss function expressed as: $\mathcal{L}=\mathcal{L}_{cls}+C\cdot\mathcal{L}_{ctr}$, where $\mathcal{L}_{cls}=\sum_{i=1}^5-\hat{y}_i\log p(y_i|\mathbf{x},\theta)$, and $\mathcal{L}_{ctr}=-logsigmoid(FC(h_{\theta}(\mathbf{x}_1))-FC(h_{\theta}(\mathbf{x}_2)))$.
Here, $C$ controls the ratio between the two losses.
For the cross-entropy loss, $\theta$ represents the model parameters of the LLM, $\mathbf{x}$ denotes the input document, $y_i$ is the predicted class, and $\hat{y}_i$ is the actual data label.
For the contrastive loss, $h_\theta$ extracts the last layer's hidden representation as the document representation, and $FC$ denotes the operation of the linear layer.

\section{Experiment}

\subsection{Experimental settings}

\paragraph{Training}

We use \texttt{LLaMA2-7b-chat}~\citep{touvron2023llama2} as the evaluator to be trained and use LoRA~\citep{conf/iclr/HuSWALWWC22} for efficient fine-tuning. During training, we set $C$ to 10. 

\paragraph{Dataset}

\begin{table*}[t]
  \centering
  \setlength{\abovecaptionskip}{0.2cm}
  \setlength{\belowcaptionskip}{-0.6cm}
  \resizebox{0.85\linewidth}{!}{
    \begin{tabular}{lcccccccc}
    \toprule
          & \multicolumn{4}{c}{Web-Article Quality} & \multicolumn{4}{c}{ASAP} \\
          \cmidrule(lr){2-5} \cmidrule(lr){6-9}
          & $\rho$ & $\tau$ & QWK   & Acc.(\%) & $\rho$ & $\tau$ & QWK   & Acc.(\%) \\
    \midrule
    TTR   & -0.243 & -0.191 & -0.101 & 41.68 & -0.2  & -0.137 & -0.159 & 37.24 \\
    Self-BLEU & 0.468 & 0.369 & -     & -     & 0.612 & 0.455 & -     & - \\
    UniEval \small{(naturalness)} & 0.114 & 0.089 & 0.007 & 6.22  & 0.107 & 0.074 & 0.073 & 19.22 \\
    UniEval \small{(understandability)} & 0.109 & 0.085 & 0.010  & 6.90   & 0.126 & 0.088 & 0.074 & 16.84 \\
    UniEval \small{(fluency)} & 0.109 & 0.085 & 0.009 & 6.75  & 0.13  & 0.091 & 0.075 & 17.69 \\
    LLM-Eval {\tiny \textit{1-5}} \small{(\texttt{GPT-3.5})} & 0.244 & 0.205 & 0.065 & 8.67  & 0.352 & 0.257 & 0.271 & 36.22 \\
    AUTO-J & 0.119 & 0.105 & 0.030  & 14.70  & 0.279 & 0.225 & 0.067 & 15.82 \\
    Prometheus & 0.240  & 0.221 & 0.163 & 6.44  & 0.104 & 0.097 & 0.009 & 1.19 \\
    \midrule
    BERT-Origin  & 0.402 & 0.383 & 0.386 & 60.57 & 0.673 & 0.633 & 0.682 & 67.18 \\
    LLaMA2-Origin   & 0.508 & 0.485 & 0.461 & \textbf{66.05} & 0.692 & 0.655 & 0.692 & 67.69 \\
    LLaMA2-MOLE   & \textbf{0.535} & \textbf{0.509} & \textbf{0.497} & \textbf{66.05} & \textbf{0.712} & \textbf{0.675} & \textbf{0.716} & \textbf{70.24} \\
    \bottomrule
    \end{tabular}%
    }
  \caption{Correlation coefficients (Spearman $\rho$, Kendall-Tau $\tau$ and Quadratic Weighted Kappa) and accuracy (Acc.) between human judgments and different metrics on Web-Article Quality Evaluation (left) and ASAP dataset (right).}
  \label{tab:all}%
\end{table*}%

We choose Web-Article Quality Evaluation and Automated Student Assessment Prize\footnote{\url{https://www.kaggle.com/c/asap-aes}} (ASAP) datasets. The former consists of long web articles in Chinese with corresponding content quality grades ranging from 0 to 4, labeled by our own. The latter is widely used in the AES task. To align with the document content quality evaluation task, we only select the sets without the source essay, which includes \texttt{set1}, \texttt{set2}, \texttt{set7}, and \texttt{set8}. Their statistical information is in Appendix~\ref{sec:statistic}.

\paragraph{Metric}

We use the Spearman correlation and Kendall-Tau correlation to measure the correlation between evaluation results and human judgments. Quadratic Weighted Kappa (QWK) is also considered, which is commonly used in the AES task~\citep{conf/acl/WangJYCGG23}.

\paragraph{Baseline}

We compare our method with 8 baselines across various methods, including diversity-based, inference-based, and training-based methods.
For diversity-based methods, 
TTR~\citep{richards1987type} and Self-BLEU\footnote{We aim to show the contribution of Self-BLEU derived from the test data to relevance to human judgments.}~\citep{conf/sigir/ZhuLZGZWY18} are used to measure the diversity of a text. 
For inference-based methods, 
UniEval~\citep{conf/emnlp/Zhong0YMJLZJH22} is a unified multi-dimensional evaluator based on T5 and re-frames NLG evaluation task as Boolean Question Answering task. 
LLM-Eval~\citep{lin2023llm} proposes a unified evaluation schema prompting LLMs to conduct multi-dimension evaluation. 
Auto-J~\citep{li2023generative} is a 13B LLM-based evaluator for evaluating alignment. 
Prometheus~\citep{kim2023prometheus} is a 13B LLM-based evaluator supporting assessing any text based on a customized score rubric. 
For training-based methods,
BERT-Origin is based on BERT-base~\citep{devlin2018bert} training on original datasets with overall quality scores. 
LLaMA2-Origin is based on \texttt{LLaMA2-7b-chat}~\citep{touvron2023llama2} after supervised fine-tuning of the original datasets.

\subsection{Results and Analysis}

\paragraph{Overall Result}

The results of all evaluators on the two datasets are presented in Table~\ref{tab:all}. 
We can observe that \textbf{MOLE outperforms the current inference-based and training-based methods across all metrics, especially the correlation with human judgment}.
For the Web-Article Quality Evaluation dataset, while MOLE performs on par with the LLaMA2-Origin in terms of accuracy, it shows certain improvements in other metrics that measure consistency with human judgment. For instance, there is an approximately 0.03 increase in the QWK metric. This indicates that our proposed MOLE framework for contrastive data construction and joint learning strategy can effectively enhance the consistency between the evaluator and human judgment by generating multi-faceted learning signals. Furthermore, on the ASAP dataset, MOLE also significantly improves consistency with human judgment, demonstrating that our method can be flexibly and effectively adapted to new evaluation tasks.

\paragraph{Extreme Scenario Analysis}

\begin{figure}[t!]
  \centering
  \setlength{\abovecaptionskip}{0.2cm}
  \setlength{\belowcaptionskip}{-0.6cm}
  \includegraphics[width=0.45\textwidth]{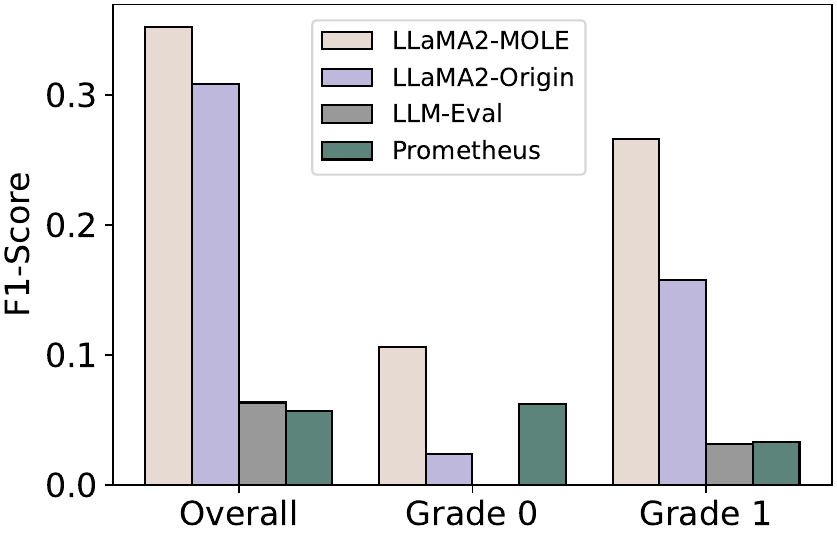}
  \caption{F1 scores for Grade 0 and Grade 1, as well as the macro F1 scores, across four different evaluators on Web-Article Quality Evaluation.}
  \label{fig:bar}
\end{figure}

In the Web-Article Quality Evaluation dataset, the F1 scores of MOLE and several representative evaluators across different levels are presented in Figure~\ref{fig:bar}. We can observe that almost all evaluators struggle with extreme cases, including Grade 0 and Grade 1, with F1 scores below 0.3. 
However, MOLE not only improves the macro F1 score but also significantly enhances the F1 score in extreme cases, particularly increasing it from 0.024 to 0.106 at Grade 0.
This demonstrates that the combination of constructed contrastive datasets with the new learning strategy enables the evaluator to better distinguish differences across varying quality levels.

\paragraph{Evaluation Capacity Analysis}

\begin{table}[t]
  \centering
  \setlength{\abovecaptionskip}{0.2cm}
  \setlength{\belowcaptionskip}{-0.6cm}
  \resizebox{\linewidth}{!}{
    \begin{tabular}{lccc}
    \toprule
          & \multicolumn{3}{c}{\texttt{set2} of ASAP} \\
\cmidrule{2-4}          & Overall & \makecell{Writing \\ Applications} & \makecell{Language \\ Conventions}  \\
    \midrule
    UniEval & 0.162 & 0.104 & 0.223 \\
    LLM-Eval & -0.057 & -0.051 & -0.047 \\
    LLaMA2-Origin & 0.719 & 0.698 & 0.658 \\
    LLaMA2-MOLE   & \textbf{0.768} & \textbf{0.755} & \textbf{0.695} \\
    \bottomrule
    \end{tabular}%
    }
  \caption{Spearman correlation coefficients between human judgments and evaluators in terms of overall and detailed scoring domains for \texttt{set2} of the ASAP dataset.}
  \label{tab:asap_set}%
\end{table}%

In the ASAP dataset, we analyze the consistency between the evaluator and human judgments across both sub-dimensions and overall dimensions. An example of the results is shown in Table~\ref{tab:asap_set}. We observe that MOLE, compared to LLaMA2-Origin, exhibited an improvement in Spearman correlation coefficients across all fine-grained dimensions by more than 0.04, demonstrating more uniform performance. This indicates that our MOLE framework, by constructing multi-faceted learning signals, effectively aids the evaluator in comprehensively perceiving various quality dimensions.

\section{Conclusion}
In this paper, we propose a framework named \textbf{M}ulti-facet c\textbf{O}unterfactual \textbf{LE}arning (MOLE), using LLMs to generate documents with counterfactual content, to construct contrastive datasets rich in multi-facet learning signals.
Furthermore, to facilitate the evaluator in fully learning the constructed signals, we propose a joint learning strategy based on contrastive learning and supervised learning. 
Experimental results demonstrate that MOLE can significantly enhance the correlation between the evaluator and human judgment, as well as rapidly adapt to new evaluation tasks.

\section*{Limitations}

There are some limitations in the MOLE framework:

\begin{itemize}
    \item The process of constructing multi-faceted counterfactual data relies on LLMs. Consequently, the capabilities of the LLM determine the quality of the generated comparative datasets, which in turn affects the evaluative capacity of the evaluators. Therefore, this process necessitates a robust LLM and incurs a certain amount of expense for data construction.
    \item The joint learning of MOLE generally provides a more even evaluative capability across various fine-grained dimensions. However, for certain complex comprehensive evaluation datasets, an increase in overall correlation with human judgments does not necessarily imply an improvement across all fine-grained dimensions.
\end{itemize}

\bibliography{custom}

\appendix

\clearpage

\section{Related Work}

The task of document content evaluation originates as a single-dimension evaluation such as coherence~\citep{conf/aaai/LiaoXLW21, conf/acl/LiuF023, conf/emnlp/MaimonT23} and readability~\citep{conf/emnlp/LiWW22, conf/sigir/LiuJYWCCG23, conf/aaai/ZengTYXH24}, or a scenario-specific task, e.g. Automated Essay Scoring (AES) task~\citep{conf/coling/XieCKZQ22, conf/acl/ChenL23, conf/acl/ChiangL23}.
Recently, many NLG evaluation efforts start to use LLMs as well-aligned evaluators, utilizing either a gradient-free approach~\citep{conf/acl/ChiangL23, li2023prd, lin2023llm} or a fine-tuned approach to evaluating the generated text~\citep{conf/emnlp/XuWPSFWL23, kim2023prometheus, wang2023pandalm}.
However, almost all existing methods for enhancing evaluators rely on a single learning signal. Given that the evaluation of document content quality is multi-faceted and highly scenario-dependent, these methods are insufficient to adequately address this evaluation task.

\section{Further Experimental Details}

\subsection{Statistic Information of Datasets}
\label{sec:statistic}
The concrete statistic of the Web-Article Quality Evaluation dataset is shown in Table~\ref{tab:article}, while the concrete statistic of the partial ASAP dataset is shown in Table~\ref{tab:asap}.

\begin{table}[htbp]
  \centering
    \begin{tabular}{cccc}
    \toprule
    Split & \#Articles & Max Len. & Avg Len. \\
    \midrule
    Train & 8660  & 1975  & 1373 \\
    Test  & 5334  & 1975  & 1468 \\
    \bottomrule
    \end{tabular}%
  \caption{Statistic information of the Web-Article Quality Evaluation dataset.}
  \label{tab:article}%
\end{table}%

\begin{table}[htbp]
  \centering
    \begin{tabular}{cccc}
    \toprule
    Set ID & \#Essays & Avg Len. & Range \\
    \midrule
    1     & 1783  & 350   & 2-12 \\
    2     & 1800  & 350   & 1-6 \\
    7     & 1569  & 250   & 0-30 \\
    8     & 723   & 650   & 0-60 \\
    \bottomrule
    \end{tabular}%
  \caption{Statistic information of the partial ASAP dataset.}
  \label{tab:asap}%
\end{table}%

\subsection{Dimensions for Refining Content Quality}
\label{sec:dimensions}

We introduce five dimensions motivated by~\citet{conf/emnlp/Zhong0YMJLZJH22} and develop more detailed descriptions:

\begin{itemize}
    \item \texttt{Coherence}: A coherent article is characterized by a logical and organized structure, clear and concise language, smooth transitions between ideas, and a seamless flow of information, ensuring that readers can easily follow and comprehend the content.
    \item \texttt{Usefulness}: An article is considered useful when it provides reliable, well-researched information, presents a comprehensive and balanced perspective, addresses relevant issues or questions, and offers practical insights or solutions for its intended audience.
    \item \texttt{Creativeness}: An article is considered creative when it demonstrates originality in its approach, offering unique perspectives, innovative ideas, and engaging storytelling that captivates and inspires the reader.
    \item \texttt{Informativeness}: An informative article is characterized by its ability to provide accurate, well-researched, and relevant information in a clear and engaging manner, catering to the needs of its target audience.
    \item \texttt{Engagingness}: Engaging articles captivate readers through a compelling combination of well-researched and relevant content, a clear and coherent structure, an accessible writing style, and the incorporation of multimedia elements that enhance understanding and maintain reader interest.
\end{itemize}

\subsection{The Data Format for Dialogue-based Metrics}
\label{sec:format}

\begin{quote}
\emph{\textbf{Question}: You are asked to write an article of high content quality for a given title, ensuring that it is coherent, useful, creative, informative and engaging.\\
Title: \{title\}\\
\textbf{Answer}: \{article\_to\_be\_evaluated\}}
\end{quote}

\clearpage
\onecolumn

\subsection{Prompts for Generating Contrastive Data}
\label{sec:prompts}

\begin{figure*}[h]
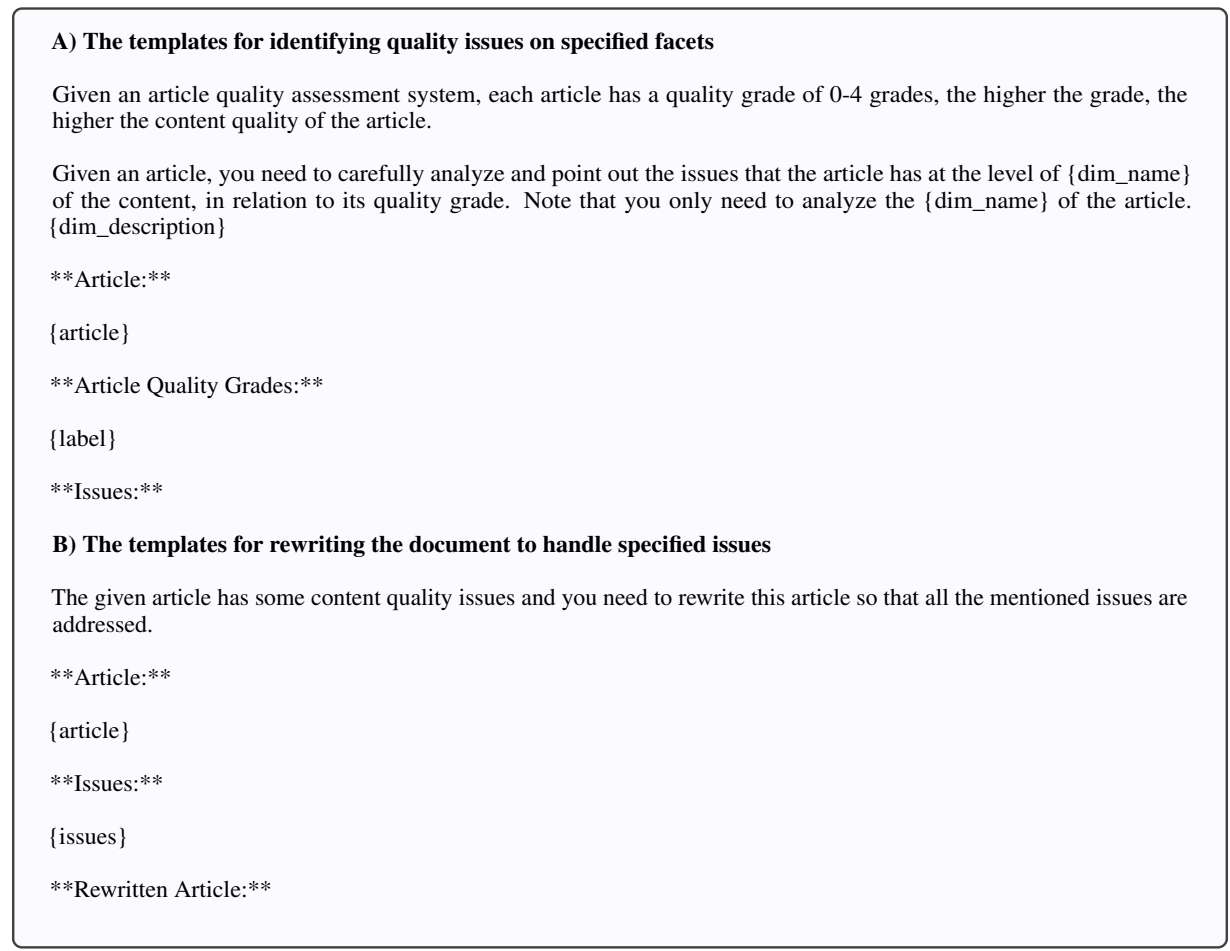

\centering
\begin{tcolorbox}[width=1\textwidth, fontupper=\small, colback=blue!2, boxrule=0.9pt] 
\textbf{A) The templates for identifying quality issues on specified facets} \\

Given an article quality assessment system, each article has a quality grade of 0-4 grades, the higher the grade, the higher the content quality of the article. \\

Given an article, you need to carefully analyze and point out the issues that the article has at the level of \{dim\_name\} of the content, in relation to its quality grade. Note that you only need to analyze the \{dim\_name\} of the article. \{dim\_description\} \\

**Article:** \\

\{article\} \\

**Article Quality Grades:** \\

\{label\} \\

**Issues:** \\

\textbf{B) The templates for rewriting the document to handle specified issues} \\

The given article has some content quality issues and you need to rewrite this article so that all the mentioned issues are addressed. \\

**Article:** \\

\{article\} \\

**Issues:** \\

\{issues\} \\

**Rewritten Article:** \\

\end{tcolorbox}
\caption{The prompt templates for generating contrastive data with counterfactual content.}
\label{fig:prompts}
\end{figure*}

The detailed prompt templates for multi-faceted counterfactual data generation (Section~\ref{sec:MFCL_1}) are shown in Figure~\ref{fig:prompts}.

\end{document}